\documentclass[11pt]{article}

\usepackage[]{EMNLP2022}

\usepackage{times}
\usepackage{latexsym}
\usepackage{pgfplots}
\usepackage[T1]{fontenc}
\usepackage[utf8]{inputenc}

\usepackage{microtype}
\usepackage{times}
\usepackage{url}
\usepackage{latexsym}

\usepackage{graphicx}
\graphicspath{ {./} }
\usepackage[table,xcdraw]{}
\usepackage{microtype}
\usepackage{booktabs}
\usepackage{amsmath}
\usepackage{color}
\usepackage{amssymb}
\usepackage{multirow} %合并多行单元格的宏包
\usepackage{makecell}
\usepackage{tabularx}
\usepackage{threeparttable}
\usepackage{amssymb}
\usepackage{marvosym}
\usepackage{subfigure}
\usepackage{soul}
\usepackage{times}
\usepackage{latexsym}

\usepackage{inconsolata}

\newcommand{\argi[1]}{\textit{#1}}
\newcommand{\argii[1]}{\textbf{#1}}
\title{Facilitating Contrastive Learning of Discourse Relational Senses by Exploiting the Hierarchy of Sense Relations}

\author{Wanqiu Long\footnotemark[2], \and Bonnie Webber\footnotemark[2] \  \\
  \footnotemark[2]\ \  University of Edinburgh, Edinburgh, UK\\
  \ \tt {Wanqiu.long@ed.ac.uk, Webber.Bonnie@ed.ac.uk}
  }
\pgfplotsset{compat=1.14}
\begin{document}
\maketitle
\begin{abstract}
Implicit discourse relation recognition is a challenging task that involves
identifying the sense or senses that hold between two adjacent spans of text, in the absence of an explicit connective between them. In both PDTB-2
\cite{prasad-etal-2008-penn} and PDTB-3 \cite{webber2019penn}, discourse relational senses are organized into a three-level hierarchy ranging from four broad top-level senses, to more specific senses below them.
Most previous work on implicit discourse relation recognition have used the sense hierarchy simply to indicate what sense labels were available. Here we do more --- incorporating the sense hierarchy into the recognition process itself and using it to select the negative examples used in contrastive learning.  With no additional effort, the approach achieves state-of-the-art performance on the task. Our code is released in
https://github.com/wanqiulong 0923/Contrastive\_IDRR.

% Most existing methods train multiple models to predict multi-level labels inde- pendently, while ignoring the dependence between hierarchically structured labels.
% We tackle implicit discourse relation classification, a task of automatically determining semantic relationships between arguments. The attention-worthy words in arguments are crucial clues for classifying the discourse relations. Attention mechanisms have been proven effective in high- lighting the attention-worthy words during encoding. However, our survey shows that some inessential words are unintentionally misjudged as the attention-worthy words and, therefore, as- signed heavier attention weights than should be. We propose a penalty-based loss re-estimation method to regulate the attention learning process, integrating penalty coefficients into the compu- tation of loss by means of overstability of attention weight distributions. We conduct experiments on the Penn Discourse TreeBank (PDTB) corpus. The test results show that our loss re-estimation method leads to substantial improvements for a variety of attention mechanism
\end{abstract}

\section{Introduction}
Discourse relations are an important aspect of textual coherence. In some cases, a speaker or writer signals the sense or senses that hold between clauses and/or sentences in a text using an explicit connective. Recognizing the sense or senses that hold can be more difficult, in the absense of an explicit connective.

% Here is an example of implicit relation:

% \begin{enumerate}

% \item[(1)] [Sears faces an especially daunting challenge on the eve of the Christmas shopping season]$_{1}$, \textbf{Implicit = because} [everyday pricing in the current environment doesn't work]$_{2}$. 

% Annotated sense: contingency.cause.reason 
% \end{enumerate}

Automatically identifying the sense or senses that hold between sentences
and/or clauses can be useful for downstream NLP tasks such as text summarization \cite{cohan-etal-2018-discourse}, machine translation \cite{article-machine-translation} and event relation extraction \cite{tang-etal-2021-discourse}. Recent studies on implicit discourse relation recognition have shown great success. Especially, pre-trained neural language models \cite{peters-etal-2018-deep, devlin-etal-2019-bert,DBLP:journals/corr/abs-1907-11692} have been used and dramatically improved the performances of models \cite{shi-demberg-2019-next,ijcai2020p530,Kishimoto2020AdaptingBT}. The senses available for labelling discourse relations in the PDTB-2 (and later in the PDTB-3) are arranged in a three-level hierarchy, with the most general senses at the top and more specific senses further down. In the PDTB-3, annotators could only choose senses at terminal nodes in the hierarchy -- level-2 senses for symmetric relations such as \textsc{Expansion.Equivalence} and \textsc{Temporal.Synchronous}, and level-3 senses for asymmetric relations, with the direction of the relation encoded in its sense label such as \textsc{Substitution.Arg1-as-subst} (where the text labelled \textsc{Arg1} substitutes for the denied text labelled \textsc{Arg2}) and \textsc{Substitution.Arg2-as-subst} (where the text labelled \textsc{Arg2} substitutes for the denied text labelled \textsc{Arg1}). Early work on recognizing the implicit relations only used the hierarchy to choose a target for recognition (e.g., the senses at level-1 (classes) or those at level-2 (types). Recently, \citet{Wu2022ALD} have tried to leverage the dependence between the level-1 and level-2 labels (cf. Section 2). The current work goes further, using the whole three-level sense hierarchy to select the negative examples for contrastive learning.
% Despite their success, these studies do not incorporate the sense hierarchy in Penn Tree bank into the recognition process, only using it to target the goal of recognition (e.g., just Level-1 senses, or just Level-2 senses). Both PDTB-2 \cite{prasad-etal-2008-penn} and PDTB-3 \cite{webber2019penn} provide sense annotations for the Explicit, Implicit and AltLex relations. As indicated in Figure 1, the set of annotated senses is organized hierarchically, with three levels: class, type and subtype. Recently,
% \citet{DBLP:journals/corr/abs-2112-11740} have tried to leverage the dependence between the hierarchically structured labels, but they mainly focus on the label dependence between level-1 and level-2. 

% Moreover, they only apply their method on PDTB-2 not on PDTB-3, while PDTB-3 make great changes at level-2 and level-3 in the sense hierarchy. The effectiveness of their method are not verified on PDTB-3. Besides, their model consists of a label attentive encoder, a label sequence decoder and an auxiliary decoder, which involves a large number of parameters. 

\begin{figure}
\setlength{\belowcaptionskip}{-0.5cm} 

    \centering
    \includegraphics[width=7.5cm]{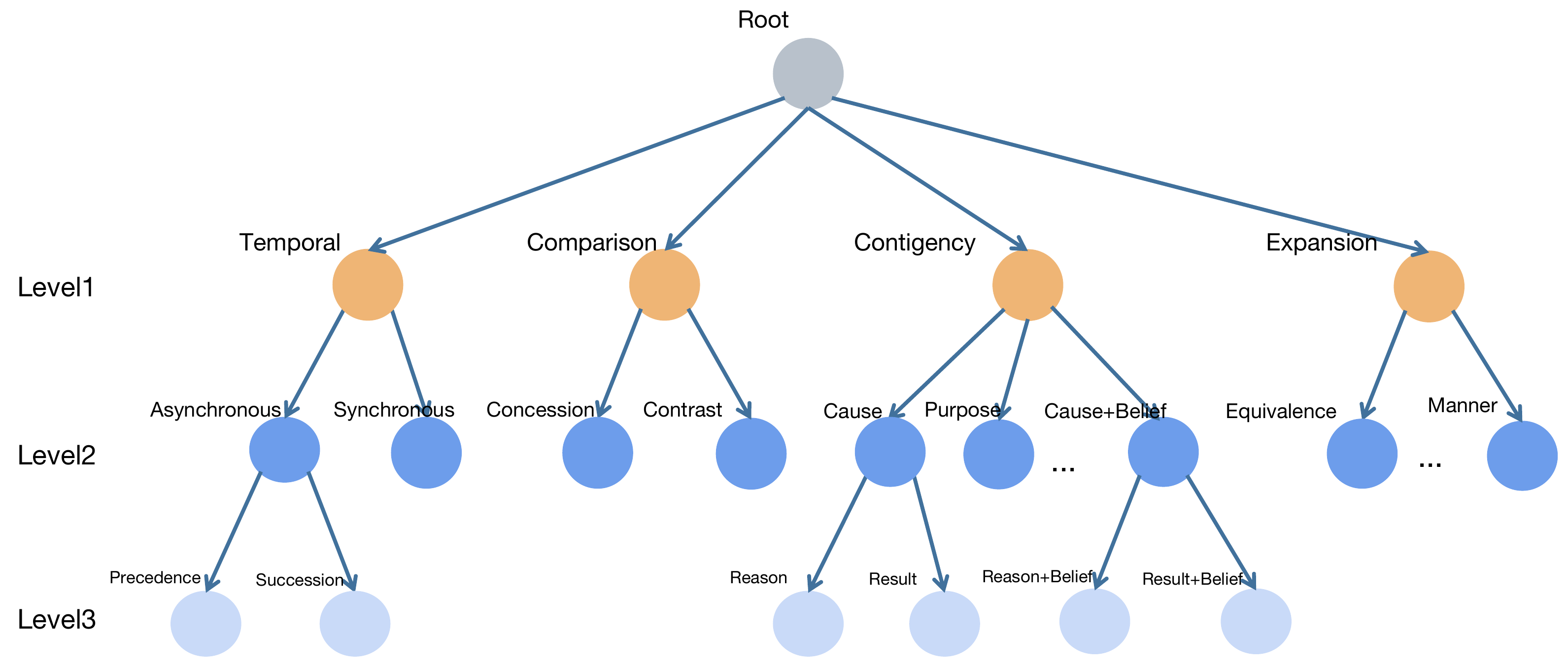}

    \caption{The PDTB-3 Sense Hierarchy}

    \label{f1}
\end{figure}

% Questions to be answered (both abstract and introduction):

% why contrastive learning for IDRR?
% why negative samples are important?
% why sense hierarchy could help?

% From your narrative:

% we learned that IDRR is crucial yet challenging, and that previous studies applied pretrained models to achieve great performance. then what motivates you to use contrastive learning for this task? what sort of problem contrastive learning could help while the other methods can't? making this clear is important to convince the readers.

% then in contrastive learning, we need to construct negative samples? why is it important? what motivates you to use sense hierarchy for constructing negative samples? you need to explain to the readers what's the sense hierarchy and the negative samples selected by sense hierarchy has special characteristics that could benefit the modeling/representation learning.

Contrastive learning, which aims to minimize the distance between similar instances (defined as positive examples) and widen the difference with dissimilar instances (negative examples), has been considered as effective in constructing meaningful representations \cite{kim-etal-2021-self, zhang-etal-2021-pairwise,yan-etal-2021-consert}. Previous work on contrastive learning indicates that it is critical to select good negative samples \cite{Alzantot2018GeneratingNL, Wu2020CLEARCL, wang-etal-2021-cline}. The insight underlying the current work is that the hierarchy of sense labels can enable the selection of good negative examples for contrastive learning. To see this, consider Examples 1-3 below from the PDTB-3. On the surface, they look somewhat similar, but in Examples 1 and 2, the annotators took the second sentence (Arg2) as providing more detail about the first sentence (Arg1) --- the sense called \textsc{Expansion.Level-of-detail.Arg2-as-detail}, while in Example 3, they took the second sentence as expressing a substitute for  ``American culture'' in terms of what is relevant -- the sense called \textsc{Expansion.Substitution.Arg2-as-subst}.

\begin{enumerate}

\item[(1)] \argi[``Valley National ''``isn't out of the woods yet '']. \argii[The key will be whether Arizona real estate turns around or at least stabilizes.]. 

\end{enumerate}

\begin{enumerate}

\item[(2)] \argi[The House appears reluctant to join the senators]. \argii[A key is whether House Republicans are willing to acquiesce to their Senate colleagues' decision to drop many pet provisions.]. 

\end{enumerate}

\begin{enumerate}

\item[(3)] \argi[Japanese culture vs. American culture is irrelevant]. \argii[The key is how a manager from one culture can motivate employees from another.].

\end{enumerate}

In this work, we use a multi-task learning framework, which consists of classification tasks and a contrastive learning task. Unlike most previous work using one benchmark dataset (usually PDTB-2 or PDTB-3), we evaluate our systems on both PDTB-2 and PDTB-3. Besides, \citet{wang-etal-2021-cline}  have shown that data augmentation can make representations be more robust, thereby enriching the data used in training. We thus follow \citet{ye-etal-2021-efficient} and \citet{NEURIPS2020_d89a66c7} in identifying a relevant form of data augmentation for our contrastive learning approach to implicit relation recognition.

The main contributions of our work are as follows:
\begin{itemize}
\setlength{\itemsep}{0pt}
\setlength{\parsep}{0pt}
\setlength{\parskip}{0pt}

    \item  We leveraged the sense hierarchy to get contrastive learning representation, learning an embedding space in which examples from same types at level-2 or level-3 stay close to each other while sister types are far apart.
    \item We explored and compared different methods of defining the negatives based on the sense hierarchies in PDTB-2 and PDTB-3, finding the approach which leads to the greatest improvements.
    \item Our proposed data augmentation method to generate examples is helpful to improve the overall performance of our model. 
    \item We demonstrate that implicit relation recognition can benefit from a deeper understanding of the sense labels and their organization. 

\end{itemize}

\section{Related Work}

\paragraph{Implicit discourse relation recognition}
For this task, \citet{dai-huang-2018-improving} considered paragraph-level context and inter-paragraph dependency. Recently, \citet{shi-demberg-2019-next} showed that using the bidirectional encoder representation from BERT \cite{devlin-etal-2019-bert} is more accurately to recognize Temporal.Synchrony, Comparison.Contrast, Expansion.Conjunction and Expansion.Alternative. \citet{ijcai2020p530} showed that different levels of representation learning are all important to implicit relation recognition, and they combined three modules to better integrate context information, the interaction between two arguments and to understand the text in depth. However, only two existing works leveraged the hierarchy in implicit relation recognition. Both \citet{Wu2020HierarchicalML} and \citet{Wu2022ALD} first attempted to assign a Level-1 sense that holds between arguments, and then only considered as possible Level-2 senses, those that are daughters of the Level-1 sense.
 
% The recently proposed pre-trained language models lead a great progress for this task. \citet{shi-demberg-2019-next} show that using the bidirectional encoder representation from BERT  \cite{devlin-etal-2019-bert} is effective to capture what events are expected to cause or follow each other. \citet{Liu2020OnTI} propose BMGF-RoBERTa (Bilateral Matching and Gated Fusion with RoBERTa), which outperforms BERT and other state-of-the-art systems on the standard PDTB dataset around 8\% and CoNLL datasets around 16\%. 

% Our work is also related to recent efforts on multi-label
% text classification, where the dependence between labels are
% used to improve the encoder for better feature extraction
% (Xie and Xing 2018; Huang et al. 2019; Zhou et al. 2020),
% or the decoder for joint label predictions (Yang et al. 2018;
% Wu, Xiong, and Wang 2019). Ours significantly differs from
% the above work in following aspects: 1) Both the encoder
% and decoder in our model can effectively leverage the label
% dependence; 2) A bottom-up auxiliary decoder is introduced
% to boost our top-down decoder via mutual learning.
\noindent
\paragraph{Contrastive learning}
% Contrastive learning aims to minimize the distance between similar instances (defined as positive examples) and widen the difference with dissimilar instances(negative examples).

Recently, there has been a growing interest in applying contrastive learning in both the pre-training and fine-tuning objectives of pre-trained language models. \citet{gao-etal-2021-simcse} used a contrastive objective to fine-tune pre-trained language models to obtain sentence embeddings, and greatly improves state-of-the-art sentence embeddings on semantic textual similarity tasks. \citet{Suresh2021NotAN} proposed label-aware contrastive loss in the presence of larger number and/or more confusable classes, and helps models to produce more differentiated output distributions. Besides, many works have demonstrated that selecting good negative examples are very important for using contrastive learning \cite{Schroff2015FaceNetAU,robinson2020hard,Cao2022ExploringTI}. In our work, we integrate contrastive learning loss with supervised losses and we use the structure of the sense hierarchy to guide the selection of negative examples.

% Contrastive learning is widely used in unsupervised or self-supervised learning \citep{MomentumContrast,Chen2020ASF,Khosla2020SupervisedCL}.

\begin{figure*}
\setlength{\belowcaptionskip}{-0.5cm} 

    \centering
    \includegraphics[width=16cm]{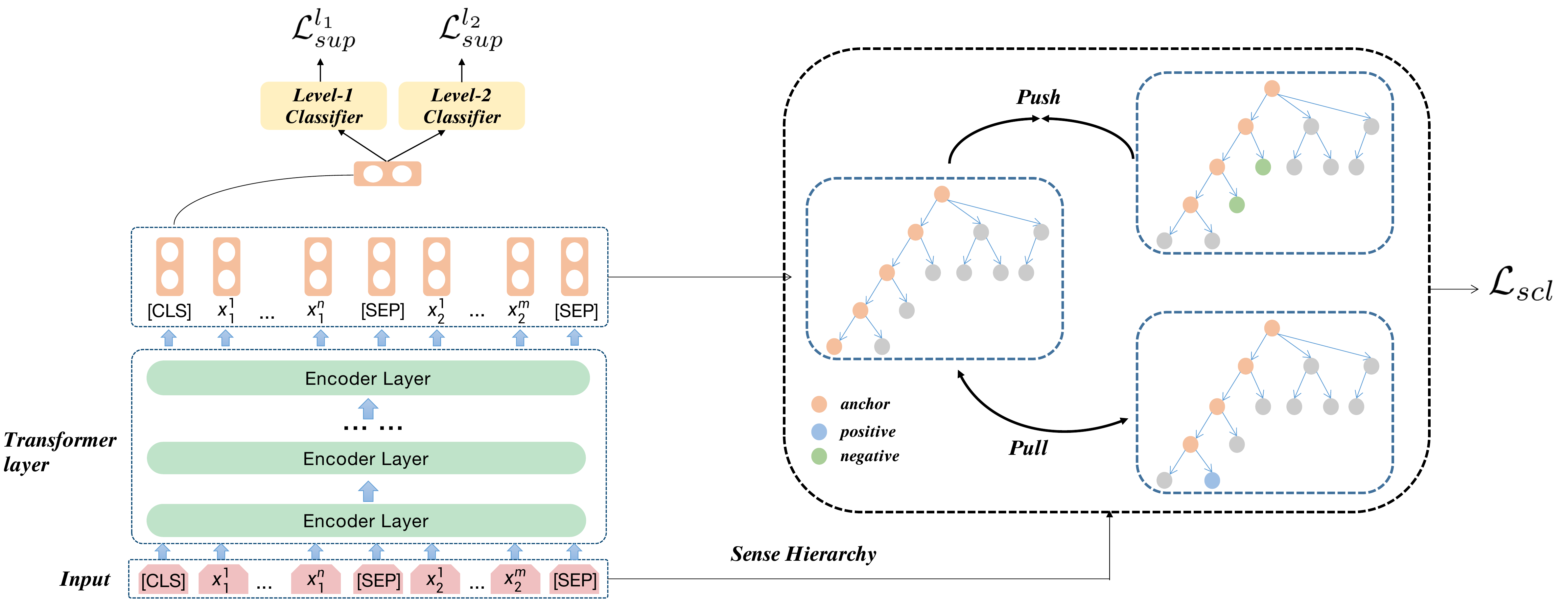}
    \caption{\textbf{The overall architecture of our model.} When given an anchor, we search the positive and negative examples in a training batch based on the sense hierarchy of the PDTB. We narrow the distances among examples from the same types at level-2 or level-3 and enlarge the distances among examples from different types at level-2 and level-3.}
    \label{fig:my_label}
\end{figure*}

%\subsection{Model Overview}
% Figure 1 shows the architecture of our model. Our model consists of six parts, including (1) a hybrid representation layer that maps each word to a hybrid between character- and word-level embedding, (2) a contextualized represen- tation layer that enhances the representation power of em- beddings, (3) a matching layer that compares each token of one argument against all tokens of the other one and vice versa, (4) a fusion layer that assigns different importance to each word in arguments based on arguments themselves as well as matching results, (5) an aggregation layer that ag- gregates fusion results and encodes each argument into a vector, (6) a prediction layer that evaluates the probability distribution P(y|Arg1, Arg2). Finally, we name our model as BMGF-RoBERTa (Bilateral Matching and Gated Fusion with RoBERTa).

%\subsection{Hierarchical Contrastive Loss}

% Here, we get loss $\mathcal{L}_{sup}^{L1}$ for the first level classification and loss $\mathcal{L}_{sup}^{L2}$ for the second level classification.

\section{Learning Loss}
\subsection{Supervised Learning Loss}
% Classifications are done in the first level and second level respectively for the same inputs. 
The standard approach today for classification task is to use a standard cross-entropy loss:
\begin{equation}
    \mathcal{L}_{sup}=\frac{1}{N}\sum_{i=1}^N-log\frac{e^{W_{y_i}^Ts_i}}{\sum_je^{W_j^Ts_i}}
\end{equation}
Where $N$ denotes the number of training examples, $y_i$ is the ground-truth class of the $i$-th class and  $W_j$ is the weight vector of the $j$-th class. 

\subsection{Contrastive Learning Loss}
In contrastive learning, each example can be treated as an anchor to get its positive and negative examples. Contrastive learning can pull the anchor and its positive example together in the embedding space, while the anchor and negative samples are pushed apart. The contrastive learning loss was used by \citet{pmlr-v119-chen20j,Suresh2021NotAN} before. A set of N randomly sampled label pairs is defined as ${x_k, y_k}$, where $x$ and $y$ represent samples and labels, respectively, $k = 1,...,N$. 
% Two increments are applied to each sample.
Let $i$ be
the index of anchor sample and $j$ is the index of a positive sample.
where $i\epsilon\{ 1,...,N\},i\ne j $.
Contrastive loss is defined as:
\begin{equation}
    \mathcal{L}_{scl}=-\sum_{i=1}^{N}\frac{e^{sim(h_j,h_i)}\tau}{\sum_{i\ne k}e^{sim(h_k,h_i)}\tau}
\label{eq2}
\end{equation}

Here, $h$ denotes the feature vector in the embedding
space, and $\tau$ is the temperature parameter. Intuitively, the numerator computes the inner dot product between the anchor points
$i$ and its positive sample $j$. The denominator computes the inner dot product between all
$i$ and the inner dot product between all negative samples.
where a total of $N - 1$ samples are computed.

Supervised contrastive learning \cite{gunel2021supervised} extends the equation.\ref{eq2} to the supervised scenario. In particular, given the presence of labels, the positive examples are all examples with the same label. The loss is defined as:
 \begin{equation}
  \begin{split}
    \mathcal{L}_{scl}=\sum_{i=1}^N-\frac{1}{N_{y_i}-1}\sum_{j=1}^{N}1_{i\ne j}1_{y_i=y_j}\\ log\frac{e^{sim(h_j,h_i)\tau}}{\sum_{k=1}^{N}1_{i\ne k}e^{sim(h_k,h_i)/\tau}}
  \end{split}
  \label{eq3}
 \end{equation}

$N_{y_j}$indicates the number of examples in a batch that have the same label as $i$, $\tau $ is the temperature parameter and $h$ denotes the feature vector that is from the l2 normalized final encoder hidden layer before the softmax projection.

\section{Our Approach}
Figure 2 shows the overall architecture of our method. As figure 2 illustrates, we firstly use a simple multi-task model based on RoBERTa-base \cite{DBLP:journals/corr/abs-1907-11692}, and then we develop a contrastive learning algorithm where the sense hierarchy is used to select positive and negative examples. Detailed descriptions of our framework and our data augmentation method are given below.

\subsection{Sentence Encoder}
Every annotated discourse relation consists of two sentences or clauses (its \textit{arguments}) and one or more relational senses that the arguments bear to each other. We concatenate the two arguments of each example and input them into RoBERTa. Following standard practices, we add two special tokens to mark the beginning ([CLS]) and the end ([SEP]) of sentences. We use the representation of [CLS] in the last layer as the representation of the whole sentences. 

% A relation may also have an explicit \textit{discourse connective} (i.e., a word or phrase that expresses the sense being conveyed). 

% In addition, as can be seen from figure xx and figurexx, not all level-2 labels have sub-types in both PDTB-2 and PDTB-3. 
% For example, in The PDTB-2 Senses Hierarchy, Synchronous at level-2 has no level-3 distinctions, while there are Precedence and Succession under Asynchronous at level-2; In PDTB-3, Contrast at level-2 has no sub-types. Considering this, we merge level-2 and level-3 together, replacing the level-2 labels with their level-3 labels if there are subtypes at level 3. In this way, negative examples for Succession are precedence and sychrounous and negative examples for reason are examples with the same level-1 but not the same level-2 and level-3 labels.
% Specifically, suppose there are 4 sentences representation $h_1$, $h_2$, $h_3$, $h_4$. where $h_1$ and $h_2$ are the same class in level-2, $h_1$ and $h_3$ are the same class in level-1 but are different classes in level-2, and $h_4$ and $h_1$, $h_2$, $h_3$ are different classes in level-1. The distance between $h_1$ and $h_2$ is $d_1$, the distance between $h_1$ and $h_3$ is $d_2$, and the distance between $h_1$ and $h_4$ is $d_3$. 

\subsection{Data Augmentation} 

To increase the number of training examples, we take advantage of meta-data
recorded with each Implicit Discourse Relation in the PDTB
(cf. \cite{webber2019penn}, Section 8]). For each sense taken to hold between the arguments of that relation, annotators have recorded in the meta-data, an explicit connective that could have signalled that sense. In the past, this meta-data was used in implicit relation recognition by both \citet{Patterson2013PredictingTP} and \citet{rutherford-xue-2015-improving}. We have used it in a 
different way, shown in Figure 3, to create an additional training example
for each connective that appears in the meta-data. In the added training
example, this added connective becomes part of the second argument of the
relation (i.e., appearing after the [SEP] character)
% To help the learned representation to be more robust and introduce much more data for training, we proposed a data augmentation method. In annotating implicit relations in both the PDTB-2 and PDTB-3, annotators were asked to identify an explicit connective for each sense that would convey each implicit sense they took to hold. This information has been retained in the data structure used to represent the corpus, as noted (for example) in Section 8 of the PDTB-3 Annotation Manual \cite{webber2019penn}.  \citet{Patterson2013PredictingTP,rutherford-xue-2015-improving} have used this information. We added the inserted connectives between two arguments to augment the data. As is shown in the figure 3 (with the connectives), the original one and the augmented one have the same label.

Since there is at least \textbf{one} explicit connective recorded in the
meta-data for each implicit discourse relation and at most \textbf{two}
\footnote{This is because the PDTB only allows for one or two senses per
relation.}, for a training batch of N tokens, there will be at least another N tokens introduced by this \textit{data augmentation} method, increasing the training batch to at least 2N tokens.

\begin{figure}[htbp]
\setlength{\belowcaptionskip}{-0.3cm}

    \centering
    \includegraphics[width=8cm]{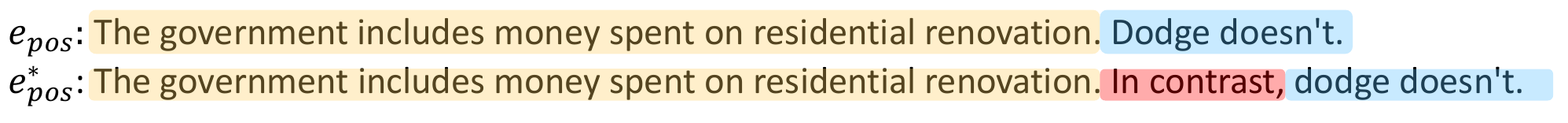}
    \caption{An example with inserted connective: the connective word is ``In contrast''.}
    \label{fig:my_label1}
\end{figure}

\subsection{Positive Pair and Negative Pair Generation}
We use the structure of the sense hierarchy to identify the positive and
negative examples needed for contrastive learning. The only senses used in annotating discourse relations are ones at terminal nodes of the sense hierarchy.  This is Level 2 for symmetric senses and Level 3 for asymmetric senses (i.e., where the \textbf{inverse} of the sense that holds between Arg1 and Arg2 is what holds between Arg2 and Arg1. For example, \textsc{Contrast}
and \textsc{Similarity} are both symmetric senses, while \textsc{Manner}
and \textsc{Condition} are asymmetric, given that there is a difference
between Arg2 being the manner of doing Arg1 or Arg1 being the manner of
doing Arg2). In our work, when the lowest level of the senses is level-3, we directly used the level-3 labels instead of their parent at level-2. For example, under the level-2 label Temporal.asynchronous, there are two labels which are precedence and succession at level-3. For this case, we replaced the level-2 label Temporal.asynchronous with the two labels precedence and succession at level-3. 

Although supervised contrastive learning in Eq. \ref{eq3} can be valid for different classes of positive example pairs, its negative examples come from any examples inside a batch except itself. We defined $l_1,l_2,l_3$ as the first, second, and third level in the hierarchical structure respectively, and $l \epsilon l_i$ refers to the labels from level $i$. 

\paragraph{Instance $e \sim$  Same sub-level} $e_{pos}$ 

\noindent Given the representation of a sentence $e_{i}$ and its first, second and third level of label $l_1^i, l_2^i, l_3^i$, we searched the set  of examples with the same second level labels or the same third level labels (if the lowest level is level-3) as $e_{pos}$ in each training batch:
\begin{gather}
    e_{pos}^{i}=\{e\in e_{pos}^{i}: l_2^e==l_2^i \quad or  \quad l_3^e==l_3^i \}
    % i_1=\{e\in i_1: l_2^e=l_2^i\} \\
    %  i_2=\{e\in i_2: l_1^e=l_1^i \quad \& \quad l_2^e\ne l_2^i\}
\end{gather}

E.g. If the label of the anchor is Temporal.asynchronous.precedence, its positive examples would be the examples with the same label. 

\paragraph{Instance $e \sim$ Batch instance  } $e_{neg}$\quad 

\noindent Here, we would like to help the model discriminate the sister types at level-2 and level-3 (if the lowest level is level-3). We searched the set of examples with different level-2 labels or level-3 labels as $e_{neg}$ in each training batch. 

E.g. If the label of the anchor is Temporal.asynchronous.precedence, its negative examples would be its sister types at level-2 and level-3, namely Temporal.asynchronous.succession and Temporal.synchronous. 

\begin{equation}
\begin{split}
    e_{neg}^{i}=\{e\in e_{neg}^{i}: l_1^e==l_1^i \quad \&  \\ (l_2^e \ne l_2^i  \quad  \&  \quad l_3^e \ne l_3^i) \}
    \end{split}
\end{equation}

\subsection{Loss Algorithms} 
As described above, given the query $e_i$ with its positive pairs and negative pairs and based on the general contrastive learning loss (see Equation \ref{eq2}), the contrastive learning loss for our task and approach is:

 \begin{equation}
  \begin{split}
    \mathcal{L}_{scl}=\sum_{i=1}^N-\frac{1}{|e^i_{pos}|-1}\sum_{j=1}^{2N}1_{i\ne j}1_{j\in e^i_{pos}}\\ log\frac{w_{j}e^{sim(h_j,h_i)\tau}}{\sum_{k=1}^{2N}1_{i\ne k}1_{k\in e_{neg}^i+e_{pos}^i}w_{k}e^{sim(h_k,h_i)/\tau}}
  \end{split}
 \end{equation}

where $w_j$ and  $w_j$ are weight factors for different positive pairs and negative pairs respectively, $sim(h_i, h_j)$ is cosine similarity and $\tau$ is a temperature hyperparameter.

%\subsection{Overall loss}
Our overall training goal is:
\begin{equation}
    \mathcal{L}=\mathcal{L}_{sup}^{l1}+\mathcal{L}_{sup}^{l2}+\beta\mathcal{L}_{scl}
    \label{eq7}
\end{equation}
As our classifications are done in the first level and second level for the same inputs, we used a standard cross-entropy loss to get supervised loss $\mathcal{L}_{sup}^{L1}$ and $\mathcal{L}_{sup}^{L2}$. And $\beta$ is the weighting factor for the contrastive loss.

\section{Experiment Setting}

\subsection{Datasets}
Besides providing a sense hierarchy, the Penn Discourse TreeBank (PDTB) also frequently serves as a dataset for evaluating the recognition of discourse relations. The earlier corpus, PDTB-2 \cite{prasad-etal-2008-penn} included 40,600 annotated relations, while the later version, PDTB-3 \cite{webber2019penn} includes an additional 13K annotations, primarily intra-sentential, as well as correcting some inconsistencies in the PDTB-2. The sense hierarchy used in the PDTB-3 differs somewhat from that used in the PDTB-2, with additions motivated by the needs of annotating intra-sentential relations and changes motivated by difficulties that annotators had in consistently using some of the senses in the PDTB-2 hierarchy.

% Penn Discourse TreeBank (PDTB) is the most frequently used benchmark dataset for evaluating the recognition of discourse relations. PDTB has two versions: PDTB-2 \cite{prasad-etal-2008-penn} and PDTB-3 \cite{webber2019penn}. PDTB-2 contains 40,600 tokens of annotated relations. 
% %  It serves as a kind of very useful resources for the development and evaluation of neural models in many downstream NLP applications \cite{narasimhan-barzilay-2015-machine,qin-etal-2017-adversarial, nie-etal-2019-dissent}. 
% Based on PDTB-2, PDTB-3 is further developed. It contains $\sim$13K more tokens annotated for discourse relations. The sense hierarchies of both PDTB-2 and PDTB-3 have three levels, with the same senses at level-1, but changes at level-2 and level-3. The figures for PDTB-2 sense hierarchy and PDTB-3 sense hierarchy can be found in Appendix. 

Because of the differences in these two hierarchies, we use the PDTB-2 hierarchy for PDTB-2 data and the PDTB-3 hierarchy for PDTB-3 data respectively. We follow earlier work \cite{Ji2015OneVI,bai-zhao-2018-deep,ijcai2020p530,xiang-etal-2022-encoding} using Sections 2-20 of the corpus for Training, Sections 0-1 for Validation, and Sections 21-22 for testing. With regard to those instances with multiple annotated labels, we also follow previous work \cite{shallow}. They are treated as separate examples during training. At test time, a prediction matching one of the gold types is taken as the correct answer. Implicit relation recognition is usually treated as a classification task. While 4-way (Level-1) classification was carried out on both PDTB-2 and PDTB-3, more detailed 11-way (Level 2) classification was done only on the
PDTB-2 and 14-way (Level 2) classification, only on the PDTB-3.

\subsection{Baselines}
To exhibit the effectiveness of our proposed method, we compare our method with strong baselines. As previous work usually used one dataset (PDTB-2 or PDTB-3) for evaluation, we use different baselines for PDTB-2 and PDTB-3. Since PDTB-3 was not released until 2019, the baselines for PDTB-3 from 2016 and 2017 are from \cite{xiang-etal-2022-encoding}. They reproduced those models which were originally used on PDTB-2 on PDTB-3. 

\noindent \textbf{Baselines for PDTB-2:}
\begin{itemize}
\setlength{\itemsep}{0pt}
\setlength{\parsep}{0pt}
\setlength{\parskip}{0pt}
\item  \cite{dai-huang-2019-regularization}: a neural model leveraging external event knowledge and coreference relations.
\item   \cite{shi-demberg-2019-learning}: a neural model that leverages the inserted connectives to learn better argument representations.
\item \cite{Nguyen2019EmployingTC}: a neural model which predicts the labels and connectives.
simultaneously.
\item  \cite{Guo2020WorkingMN}: a knowledge-enhanced Neural Network framework.
\item \cite{Kishimoto2020AdaptingBT}: a model applying three additional training tasks.
\item \cite{ijcai2020p530}: a RoBERTa-based model which consists of three different modules.
% \item \cite{DBLP:journals/corr/abs-2112-11740}: a model incorporates the input representation and its level-specific context, then it uses a decoder to output the prediction of the model's corresponding level in a top-down manner, trained with mutual learning.
\item \cite{jiang-etal-2021-just}: a method that recognizes the relation label and generates the target sentence simultaneously.
\item \cite{dou-etal-2021-cvae-based}: a method using conditional VAE to estimate the risk of erroneous sampling.
\item \cite{Wu2022ALD}: a label dependence-aware sequence generation model. 
\end{itemize}

\textbf{Baselines for PDTB-3:}
\begin{itemize}
\setlength{\itemsep}{0pt}
\setlength{\parsep}{0pt}
\setlength{\parskip}{0pt}
\item \cite{liu-li-2016-recognizing}: a model that combines two arguments’ representation for stacked interactive attention.
\item \cite{Chen2016DiscourseRD}: a mixed generative-discriminative framework.
\item \cite{lan-etal-2017-multi}: a multi-task attention neural network.
\item \cite{ruan-etal-2020-interactively}: a propagative attention learning model.
\item \cite{xiang-etal-2022-encoding}: a model that uses a Dual Attention Network (DAN).
\end{itemize}

\subsection{Parameters Setting}
In our experiments, we use the pre-trained RoBERTa-base \cite{DBLP:journals/corr/abs-1907-11692} as our Encoder. We adopt Adam \cite{kingma:adam} with the learning rate of $3e{-}5$ and the batch size of 256 to update the model. The maximum training epoch is set to 25 and the wait patience for early stopping is set to 10 for all models. We clip the gradient L2-norm with a threshold 2.0. For contrast learning, the weight of positive examples is set to 1.6 and the weight of negative examples is set to 1. All experiments are performed with 1× 80GB NVIDIA A100 GPU.
\begin{figure}
\setlength{\belowcaptionskip}{-0.3cm} 

    \centering
    \includegraphics[width=8cm]{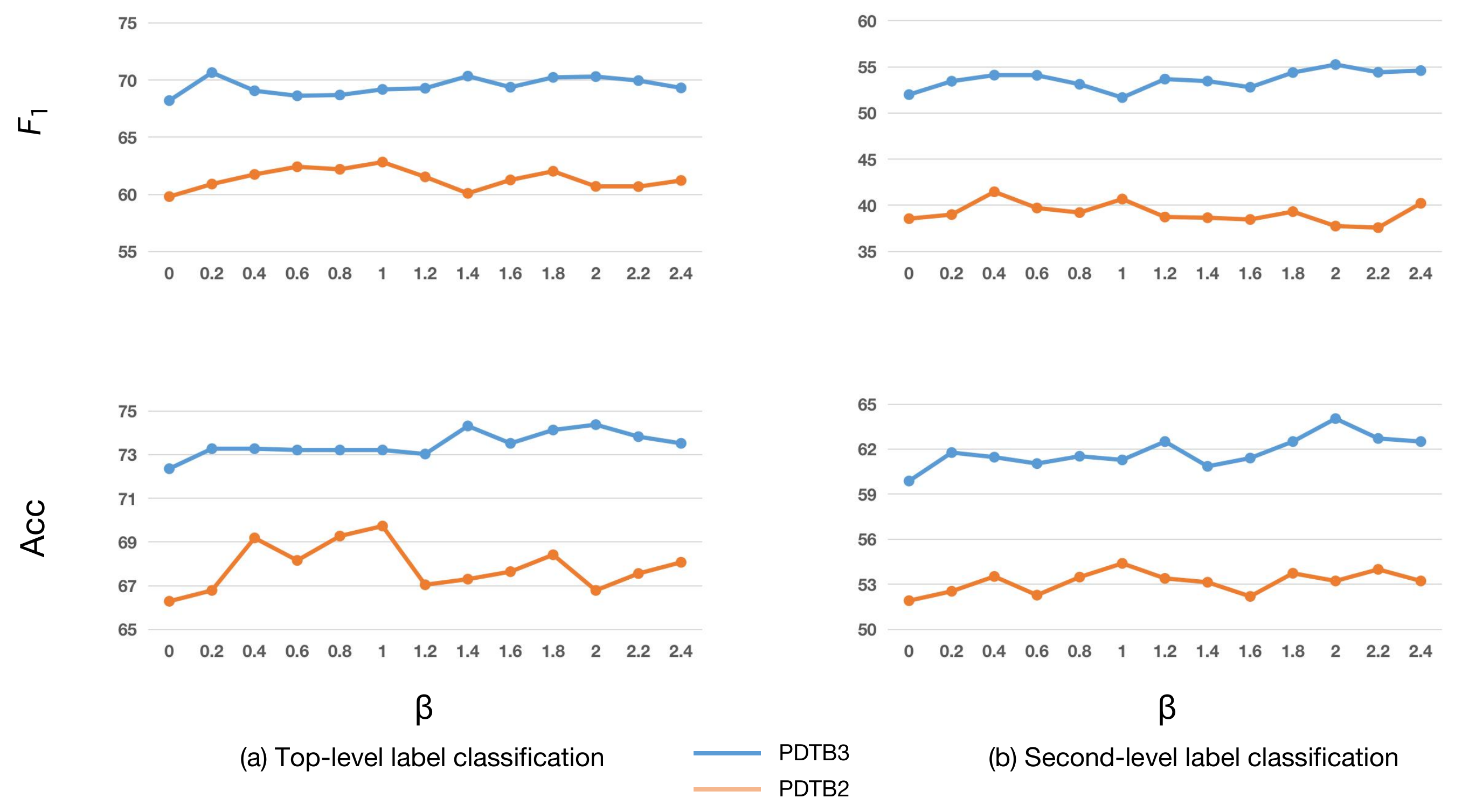}

    \caption{Effects of $\beta$ on the validation set.}

\end{figure}
\subsection{Evaluation Metrics}
We used Accuracy and Macro-F1 score as evaluation metrics, because PDTB datasets are imbalanced and Macro-F1 score has been said to be an more appropriate assessment measure for imbalanced datasets \cite{Akosa2017PredictiveA, Bekkar2013EvaluationMF}.

\subsection{Effects of the Coefficient $\beta$}
As shown in Equation \ref{eq7}, the coefficient $\beta$ is an important hyperparameter that controls the relative importance of supervised loss and contrastive loss. Thus, we vary $\beta$ from 0 to 2.4 with an increment of 0.2 each step, and inspect the performance of our model using different $\beta$ on the validation set. 

From Figure 4, we can find that, compared with the model without contrastive learning ($\beta$ = 0), the performance of our model at any level is always improved via contrastive learning. For PDTB-2, when $\beta$ exceeds 1.0, the performance of our model tends to be stable and declines finally. Thus, we directly set $\beta$ = 1.0 for all PDTB-2 related experiments thereafter. For PDTB-3, the Acc and F1 of the validation set reach the highest point at $\beta$ = 2.0. Therefore we choose $\beta$ = 2.0 for all related experiments. 

We have considered three ways of investigating why there is such a difference in the optimal weighting coefficient. First, compared with PDTB-2, the PDTB-3 contains about 6000 more implicit tokens annotated for discourse relations. Secondly, although the sense hierarchies of both the PDTB-2 and the PDTB-3 have three levels and have the same senses at level- 1, but many changes at level-2 and level-3 due to difficulties found in annotating certain senses. Moreover, the intra-sentential implicit relations might be another reason. In PDTB-3, many more discourse relations are annotated within sentences. \citet{liang-etal-2020-extending} report quite striking difference in the distribution of sense relations inter-sententially vs. intra-sententially between PDTB-2 and PDTB-3. Therefore, these major differences in the PDTB-3 and the PDTB-2 might cause the fluctuation of the coefficient value.

\section{Results and Analysis}
The results on PDTB-2 and PDTB-3 for Level-1 and Level-2 are presented in Table 1 and Table 2 respectively, where the best results are highlighted in bold. Classification performance on PDTB-2 in terms of Macro-F1 for the four general sense types at Level-1 and 11 sense types at Level-2 is shown in Table 3 and Table 4.

These results demonstrate better performance than previous systems for both Level-1 and Level-2 classification on both PDTB-2 and PDTB-3. In particular, the results clearly demonstrate benefits to be gained from contrastive learning. But there is more to be said: In Section 6.1, we discuss different ways of defining negative examples with respect to the sense hierarchy, and in Section 6.2, we discuss the relative value of the particular form of data augmentation we have used (cf. Section 4.2) as compared with our method of contrastive learning.

\begin{table}
\centering
\scriptsize
\begin{tabular}{l|cc|cc}
\bottomrule
\multirow{3}{*}{Model}&\multicolumn{4}{c}{PDTB-2} \\

&\multicolumn{2}{c|}{Top Level}&\multicolumn{2}{c}{Second Level} \\

&Acc&Macro-F1&Acc&Macro-F1\\
\hline
\citet{dai-huang-2019-regularization}&59.66&52.89&48.23&33.41\\
\citet{shi-demberg-2019-learning}&61.42&46.40&47.83&-\\
\citet{Nguyen2019EmployingTC}&-&53.00&49.95&-\\
\citet{Guo2020WorkingMN}&57.25&47.90&-&-\\

\citet{Kishimoto2020AdaptingBT}&65.26&58.48&52.34&-\\
\citet{ijcai2020p530}&69.06&63.39&58.13&-\\
\citet{jiang-etal-2021-just}&-&57.18&-&37.76\\
\citet{dou-etal-2021-cvae-based}&70.17&65.06&-&-\\
\citet{Wu2022ALD}&71.18&63.73&60.33&40.49\\

\hline
Ours&\textbf{72.18}&\textbf{69.60}&\textbf{61.69
}&\textbf{49.66}
\\
\hline
\toprule
\end{tabular}
\caption
{Experimental results on PDTB-2.}
\end{table}

\subsection{Comparisons with Other Negatives Selecting Methods}
There is not only one way to select negative examples for contrastive learning based on PDTB hierarchical structures. In addition to the method we adopt, we have explored another 4 different methods of defining positive and negative examples by using the sense hierarchies, which can be shown in Figure 5. One can choose the level against which to select negative examples: method 2 below uses examples with different labels at level-2, while methods 1, 3 and 4 use examples with different labels at level-1. With regard to the use of weight for method 3 and method 4, we aim to give more weight to more similar (potentially) positive examples based on the hierarchy. Specifically, we give more weight to the examples from the same level-2/level-3 type than their sister types at level-2/level-3 when all of the examples from the same level-1 are positive examples. Besides, method 4 leverages level-3 labels, while method 1 to 3 only consider level-1 and level-2 labels.
\begin{table}
\setlength{\belowcaptionskip}{-0.1cm}
\centering
\scriptsize
\begin{tabular}{l|cc|cc}
\bottomrule
\multirow{3}{*}{Model}&\multicolumn{4}{c}{PDTB-3} \\

&\multicolumn{2}{c|}{Top Level}&\multicolumn{2}{c}{Second Level} \\

&Acc&Macro-F1&Acc&Macro-F1\\
\hline
\citet{liu-li-2016-recognizing}&57.67&46.13&-&-\\
\citet{chen-etal-2016-implicit}&57.33&45.11&-&-\\

\citet{lan-etal-2017-multi}&57.06&47.29&-&-\\
\citet{ruan-etal-2020-interactively}&58.01&49.45&-&-\\
\citet{xiang-etal-2022-encoding} &60.45&53.14&-&-\\
(BiLSTM)&&&&\\
\citet{xiang-etal-2022-encoding}& 64.04&56.63&-&-\\
(BERT)&&&&\\
\hline
% RoBERTa&72.02&67.44&60.56&57.12\\
% RoBERTa-MTL&72.63&68.23&60.56&57.16\\
% \hline
Ours&\textbf{75.31}&\textbf{70.05}&\textbf{64.68}&\textbf{57.62}
\\
\hline
\toprule
\end{tabular}
\caption
{Experimental results on PDTB-3.}
\end{table}
\begin{table}[!tbp]
\centering
\scriptsize
\begin{tabular}{l|cc|cc}
\bottomrule

Model&Comp.&Cont&Exp.&Temp.\\
\hline
\citet{Nguyen2019EmployingTC}&48.44& 56.84 &73.66 &38.60\\
\citet{Guo2020WorkingMN}&43.92& 57.67 &73.45 &36.33\\
\citet{ijcai2020p530}&\underline{59.44} &60.98 &77.66 & \underline{50.26}\\
\citet{jiang-etal-2021-just}&55.40& 57.04 &74.76& 41.54\\
\citet{dou-etal-2021-cvae-based}&55.72 &\underline{63.39} &\textbf{80.34} &44.01\\

\hline
% RoBERTa&69.10&64.67&56.40&43.74\\
% RoBERTa-multi&67.37&62.04&55.53&36.85\\
% RoBERTa&61.39 &59.66 &75.75 &62.67 \\
% RoBERTa-MTL&60.84 &63.57 &77.31 &59.85 \\
% \hline
% $\rm RoBERTa_{base}$+$\rm SCL_{WL1}$&63.26 &60.42 &76.78 &59.74 \\
% $\rm RoBERTa_{base}$+$\rm SCL_{WL2}$&60.78 &60.82 &77.89 &56.30 \\

% $\rm RoBERTa_{base}$+$\rm SCL_{WL1+WL2}$&59.85 &65.18 &76.43 &64.67 \\
% $\rm RoBERTa_{base}$+$\rm SCL_{WL1+WL2+WL3}$&57.25&61.73&77.30 &64.90 \\
% \hline
Ours&\textbf{65.84} &\textbf{63.55} &\underline{79.17} &\textbf{69.86} 
\\
\hline
\toprule
\end{tabular}
\caption
{The results for relation types at level-1 on PDTB-2 in terms of F1 (\%) (top-level multi-class classification).}
\label{table-1}
\end{table}
\begin{table}
\setlength{\belowcaptionskip}{-10pt}
\centering
\scriptsize
\begin{tabular}{l|cc|c}
\bottomrule

Second-level Label&\citet{ijcai2020p530}& \citet{Wu2022ALD}&Ours\\
\hline
Temp.Asynchronous&56.18& 56.47 & 	\textbf{59.79}\\
Temp.Synchrony&0.00& 0.00&	\textbf{78.26} \\
\hline
Cont.Cause&59.60 &64.36 &	\textbf{65.58} \\
Cont.Pragmatic cause&0.0 &0.0&	0.00 \\
\hline
Comp.Contrast &59.75& \textbf{63.52}&	62.63 \\
Comp.Concession& 0.0 &0.0&	0.00 \\
\hline
Exp.Conjunction &\textbf{60.17}& 57.91 	&58.35 \\
Exp.Instantiation& 67.96 &72.60& 	\textbf{73.04} \\
Exp.Restatement &53.83 &58.06 &	\textbf{60.00} \\
Exp.Alternative  &60.00& \textbf{63.46}&	53.85 \\
Exp.List &0.0 &8.98 &	\textbf{34.78} \\
% \citet{Guo2020WorkingMN}&43.92& 57.67 &73.45 &36.33\\
% \citet{Nguyen2019EmployingTC}&48.44& 56.84 &73.66 &38.60\\
% \citet{Liu2020OnTI}&59.44 &60.98 &77.66 & 50.26\\
% \citet{jiang-etal-2021-just}&55.40& 57.04 &74.76& 41.54\\
% \citet{dou-etal-2021-cvae-based}&55.72 &63.39 &80.34 &44.01\\
% \citet{DBLP:journals/corr/abs-2112-11740}&\\

% \hline
% % RoBERTa&69.10&64.67&56.40&43.74\\
% % RoBERTa-multi&67.37&62.04&55.53&36.85\\
% RoBERTa&61.39 &59.66 &75.75 &62.67 \\
% RoBERTa-MTL&60.84 &63.57 &77.31 &59.85 \\
% \hline
% $\rm RoBERTa_{base}$+$\rm SCL_{WL1}$&63.26 &60.42 &76.78 &59.74 \\
% $\rm RoBERTa_{base}$+$\rm SCL_{WL2}$&60.78 &60.82 &77.89 &56.30 \\

% $\rm RoBERTa_{base}$+$\rm SCL_{WL1+WL2}$&59.85 &65.18 &76.43 &64.67 \\
% $\rm RoBERTa_{base}$+$\rm SCL_{WL1+WL2+WL3}$&57.25&61.73&77.30 &64.90 \\
% \hline
% Ours&65.84 &63.55 &79.17 &69.86 
% \\
\hline
\toprule
\end{tabular}
\caption
{The results for relation types at level-2 on PDTB-2 in terms of F1 (\%) (second-level multi-class classification).}
\label{table-4}
\end{table}
In our experiments for other negatives defining methods, we use the same hyperparameters as the experimental setup of our methods.For method 3 and method 4, the weight of positive examples is set to 1.6 and 1.3 and the weight of negative examples still is 1.
%The learning rate is $3e{-}5$ and the batch size is 256
\begin{table*}[!htbp]
\setlength{\belowcaptionskip}{-6pt}

\centering
\footnotesize
\begin{tabular}{l|cc|cc|cc|cc}
\bottomrule
\multirow{3}{*}{Model}&\multicolumn{4}{c|}{PDTB-2}&\multicolumn{4}{c}{PDTB-3} \\

&\multicolumn{2}{c|}{Top Level}&\multicolumn{2}{c|}{Second Level}&\multicolumn{2}{c|}{Top Level}&\multicolumn{2}{c}{Second Level} \\

&Acc&Macro-F1&Acc&Macro-F1&Acc&Macro-F1&Acc&Macro-F1\\
\hline
% $\rm RoBERTa_{base}$+$\rm SCL_{no}$ \cite{gunel2021supervised}&&&&&&&&\\
Method 1&68.91&65.04&58.61&46.27&73.25&68.00&61.17&55.58\\
Method 2&69.39&63.95&58.33&44.80&73.53&68.36&61.93&54.85\\

Method 3&69.39&66.53&58.61&39.20&72.49&67.49&60.77&54.33\\
%Method 3&69.39&65.88&58.52&46.29&72.49&67.49&60.77&54.33\\
Method 4&69.10&65.30&57.07&47.46&71.26&66.47&59.53&47.24\\
\hline 
Ours&\textbf{72.18}&	\textbf{69.60}	&\textbf{61.69}&	\textbf{49.66}&\textbf{75.31}&\textbf{70.05}&\textbf{64.48}&\textbf{	57.62}\\
\hline
\toprule
\end{tabular}
\caption
{Comparisons with other negatives
defining methods.}
\label{table-6}
\end{table*}

It can be seen from table 5 and table 6 that our method is better than the above methods in both datasets for both level-1 and level-2 classification tasks. Compared with method 2, we utilize level-3 labels, which indicated the level-3 label information is helpful for the approach. The greatest difference between our method and other three methods is that our negative examples are only those sister types at level-2 or level-3, not including the examples from different level-1. On the contrary, the negative examples in those three methods are examples from other level-1 types. We suppose that this might make a too strong assumption that examples from different level-1 are very dissimilar. In PDTB datasets, some examples have been annotated with multiple labels. We found that among all examples with multiple annotated labels, there are 99.26\% examples whose multiple labels are under different level-1. Moreover, some level-1 types of relation might be overlapped even if the annotators just annotate one label. For example, some examples annotated as Temporal.asynchronous might have the sense of Contingency.cause as well. And \citet{moens-steedman-1988-temporal} have pointed out that \textit{when}-clauses do not simply predicate a temporal relation, but a causal one as well, which can be called contingency. This shows up in the PDTB in terms of the variation in how particular tokens of \textit{when} clauses have been annotated.  But it also means that in choosing Negative examples, relations labelled \textsc{Temporal.Synchronous} or \textsc{Temporal.Asynchronous} may closely resemble those labelled \textsc{Contingency.Cause} and therefore not be effective as negative examples. Specifically, for the following example:
\begin{enumerate}
    \item [(4)]
\textbf{when} [they built the 39th Street bridge]$_{1}$,
[they solved most of their traffic problems]$_{2}$. 
\end{enumerate}
If the connective  ``when'' is replaced with ``because'', the sentence still sounds not strange. Therefore, regarding all examples from different level-1 as negative examples might have some negative impacts on learning the representations.
\begin{figure}[!tbp]
\setlength{\belowcaptionskip}{-8pt}

\centering
\subfigure[method 1]{
\includegraphics[width=3.3cm]{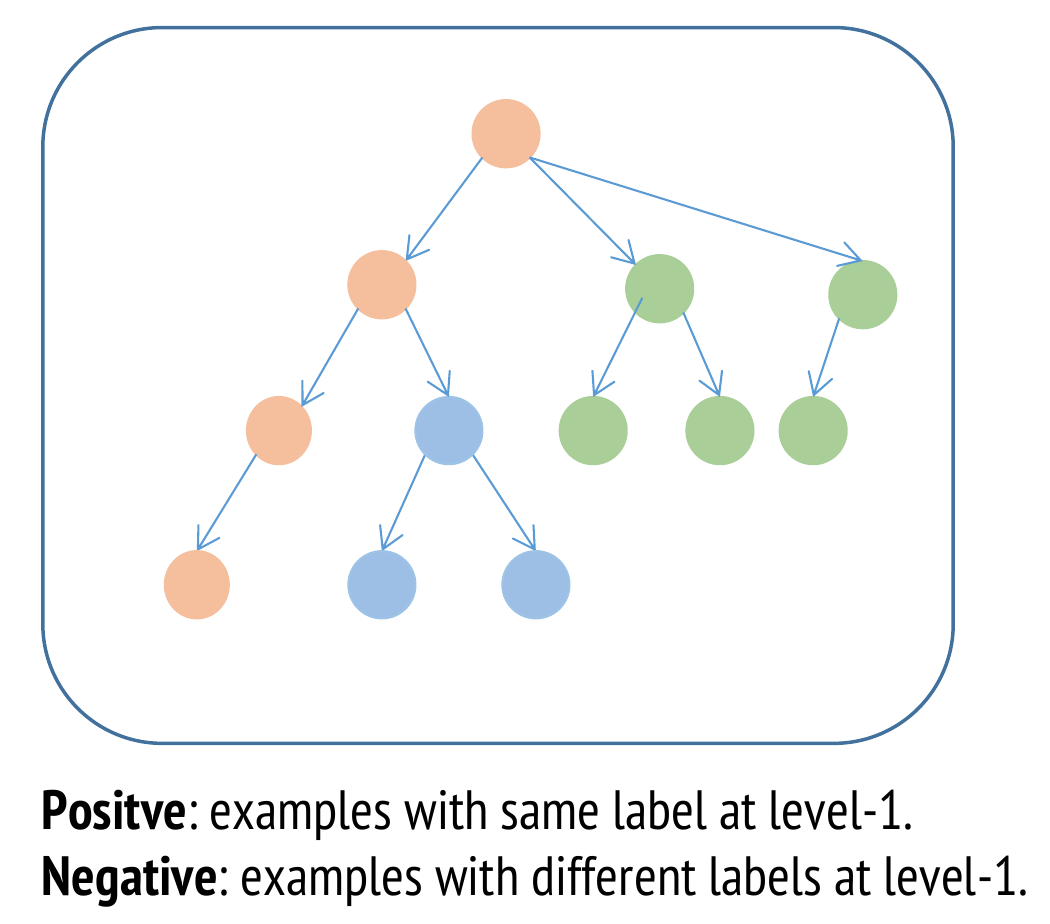}
%\caption{fig1}
}
\quad
\subfigure[method 2]{
\includegraphics[width=3.3cm]{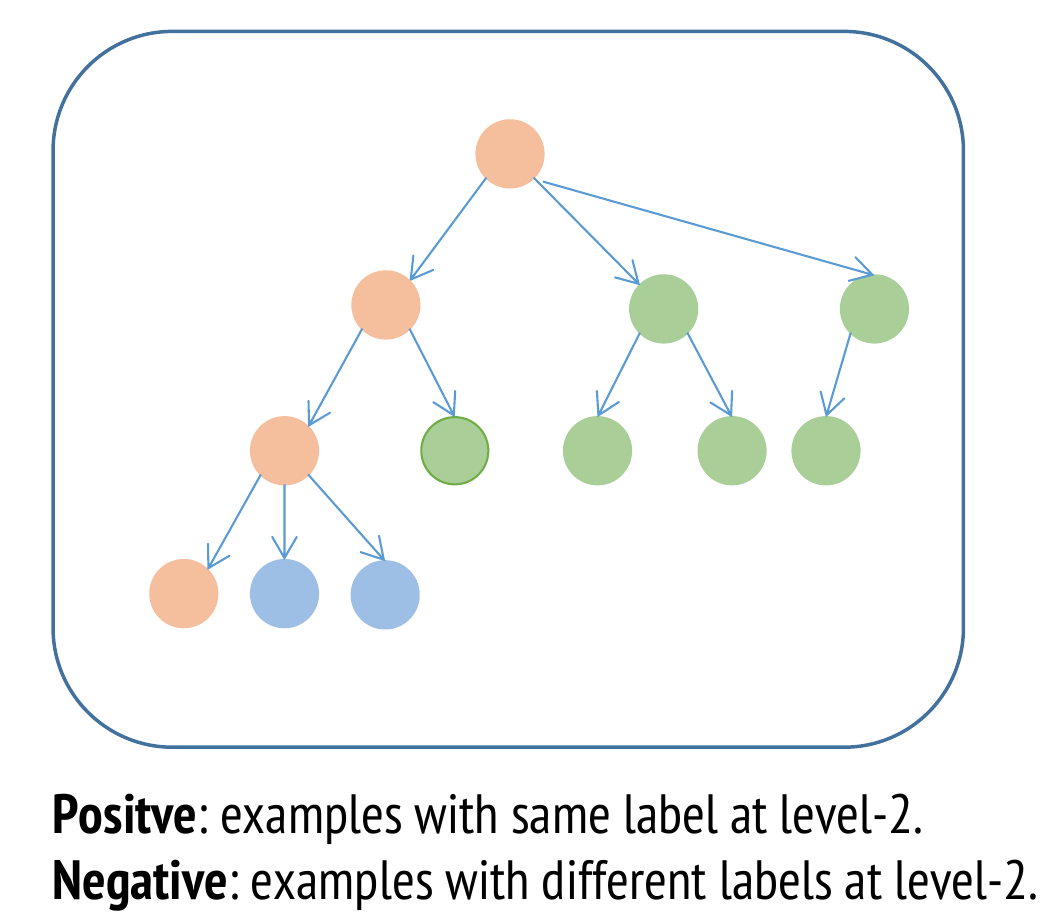}
}
\quad
\subfigure[method 3]{
\includegraphics[width=3.3cm]{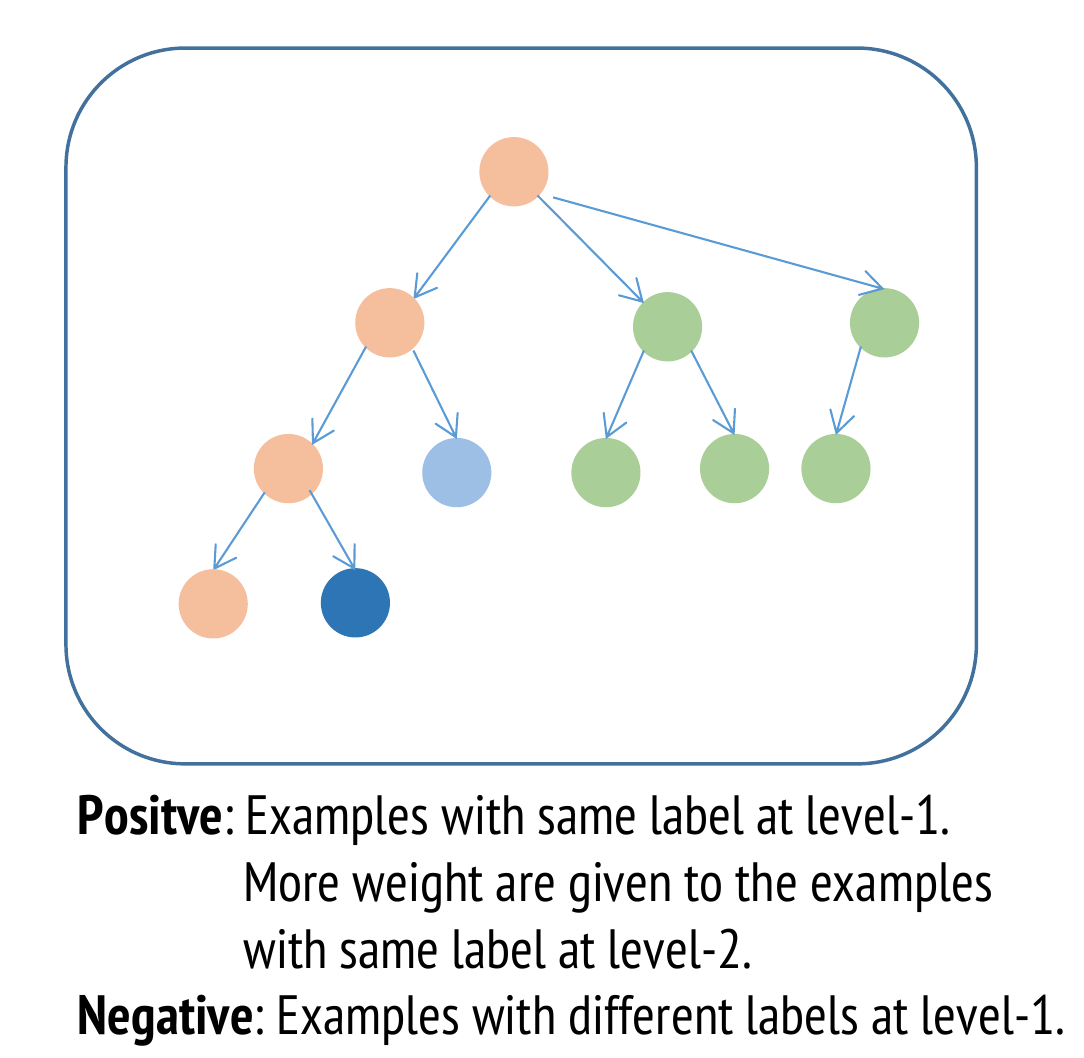}
}
\quad
\subfigure[method 4]{
\includegraphics[width=3.3cm]{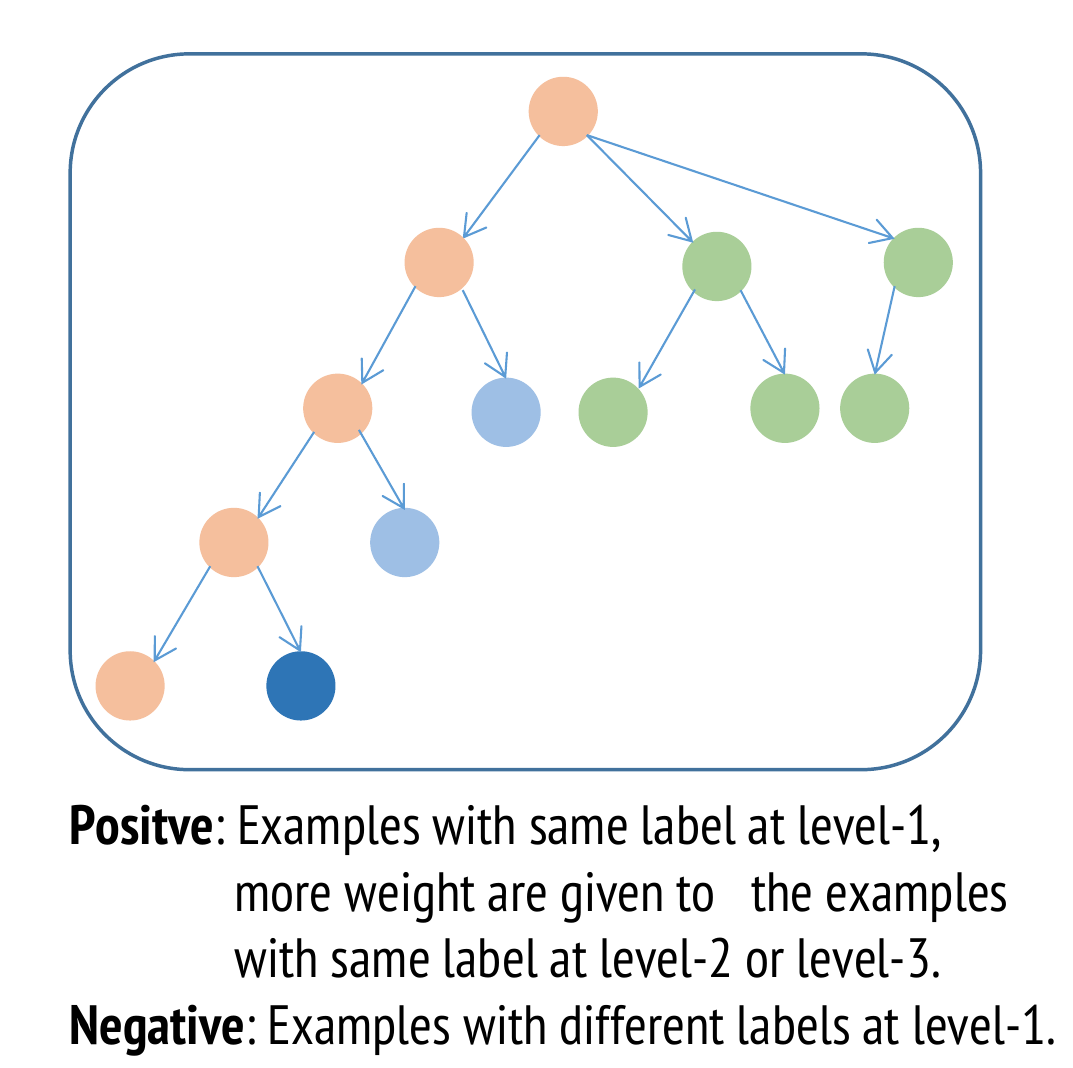}
}
\caption{Another four negative examples selected methods. \textbf{\textcolor[RGB]{245,191,157}{orange}} ball represent anchor, \textbf{\textcolor[RGB]{182,213,167}{green}} ball represent negative examples, and \textbf{\textcolor[RGB]{173,203,234}{blue}} ball represent positive examples. \textbf{\textcolor[RGB]{57,137,195}{Darker blue}} ball means more weight is given to more similar (potentially) positive examples.}
\end{figure}

\begin{table}[tbp]

\centering
\small
\begin{tabular}{l|cc|cc}
\bottomrule

Model&Comp.&Cont&Exp.&Temp.\\
\hline
Method 1&63.26 &60.42 &76.78 &59.74 \\
Method 2&60.78 &60.82 &77.89 &56.30 \\

Method 3&59.85 &\textbf{65.18} &76.43 &64.67 \\
%Method 3&61.77&62.13&77.36&62.25\\
Method 4&57.25&61.73&77.30 &64.90 \\
\hline
Ours&\textbf{65.84} &63.55 &\textbf{79.17} &\textbf{69.86 }
\\
\toprule
\end{tabular}
\caption
{The results of relation types at level-1 on PDTB-2 in terms of F1 (\%) (top-level multi-class classification).}

\end{table}
\begin{table*}[!tbp]
\centering
\footnotesize
\begin{tabular}{ll|cc|cc}
\bottomrule
\multirow{2}{*}{Datasets}&\multirow{2}{*}{Model}&\multicolumn{2}{c|}{Top Level}&\multicolumn{2}{c}{Second Level}\\ %\multicolumn{4}{c}{PDTB3} \\

% &\multicolumn{2}{c|}{Top Level}&\multicolumn{2}{c}{Second Level} \\

&&Acc&Macro-F1&Acc&Macro-F1\\
\hline

% RoBERTa&69.10&64.67&56.40&43.74\\
% RoBERTa-multi&67.37&62.04&55.53&36.85\\
\multirow{3}{*}{PDTB-2}&RoBERTa&68.14&64.87&58.33&48.37\\
&RoBERTa-MTL&69.87&65.39&58.22&45.21\\

&Ours&\textbf{72.18}&\textbf{69.60}&\textbf{61.69
}&\textbf{49.66}\\

\hline 
\multirow{3}{*}{PDTB-3}&RoBERTa&72.02&67.44&60.56&57.12\\
&RoBERTa-MTL&72.63&68.23&60.56&57.16\\
&Ours&\textbf{75.31}&\textbf{70.05}&\textbf{64.68}&\textbf{57.62}
\\
\hline
\toprule
\end{tabular}
\setlength{\belowcaptionskip}{-0.2cm}
\caption
{Ablation study on PDTB-2 and PDTB-3. }
\label{table-mtl}
\end{table*}
\begin{table}[!tbp]
\setlength{\belowcaptionskip}{-6pt}
\centering
\scriptsize
\begin{tabular}{ll|cc|cc}
\bottomrule
% \multirow{2}{*}{Model}&&\multicolumn{4}{c}{PDTB3} \\

\multirow{2}{*}{Model}&&\multicolumn{2}{c|}{Top Level}&\multicolumn{2}{c}{Second Level} \\

&&Acc&Macro-F1&Acc&Macro-F1\\
\hline
\multirow{2}{*}{PDTB-2}&Ours&\textbf{72.18}&\textbf{69.60}&\textbf{61.69
}&\textbf{49.66}
\\
%& -augmentation&68.14&64.96&56.88&47.64 \\%这里调了
%& -augmentation&70.23&67.37&59.48&47.64\\
& -augmentation&71.70&67.85&59.19&45.54\\
\hline
\multirow{2}{*}{PDTB-3}&Ours&\textbf{75.31}&\textbf{70.05}&\textbf{64.68}&\textbf{57.62}
\\
& -augmentation&73.32&69.02&63.24&51.80 \\
\hline
\toprule
\end{tabular}
\setlength{\belowcaptionskip}{-0.5cm}

\caption
{Effects of data augmentation.}
\label{table-data-aug}
\end{table}

\subsection{Ablation Study}
We wanted to know how useful our data augmentation method and our contrastive learning method are, so we have undertaken ablation studies for this.

\noindent\textbf{Effects of contrastive learning algorithm} \quad From Table \ref{table-mtl}, it can be seen that multi-task learning method where level-1 and level-2 labels are predicted simultaneously by using the same [CLS] representation perform better than separately predicting level-1 and level-2 labels, which verifies the dependency between different levels. Compared with the multi-task learning method, our model with a contrastive loss has better performance in PDTB-2 and PDTB-3, which means that our contrasting learning method is indeed helpful.

\noindent\textbf{Effects of data augmentation}\quad Table \ref{table-data-aug} compares the results with and without data augmentation for both PDTB-2 and PDTB-3. From the comparisons, it is clear that the data augmentation method is helpful to generate useful examples. \citet{NEURIPS2020_d89a66c7} showed that having a large number of hard positives/negatives in a batch leads to better performance. Since we have many classes at the second level, 11 types for PDTB-2 and 14 types for PDTB-3. In a batch with the size of 256, it is difficult to guarantee that there are enough positive examples for each class to take full advantage of contrast learning. Therefore, without data augmentation, the performance of our method degrades considerably.

% We will incorporate this suggestion and include an example such as the following:

% E.g. That isn't all. (For instance) Last year, the Irish airport authority, in a joint venture with Aeroflot, opened four hard-currency duty-free shops at Moscow's Sheremetyevo Airport.  
 

\section{Limitations and Future work}
With regard to PDTB-2 and PDTB-3 annotation, there are two cases: (1) Annotators can assign multiple labels to an example when they believe more than one relation holds simultaneously; (2) Annotators can be told (in the Annotation Manual) to give precedence to one label if they take more than one to hold. For example, they are told in the Manual \cite{webber2019penn} that examples that satisfy the conditions for both Contrast and Concession, should be labelled as concession. We over-simplified the presence of multiple labels by following \citet{qin-etal-2017-adversarial} in treating each label as a separate example and did not consider the second case. Thus, our approach might be inadequate for dealing with the actual distribution of the data and can be extended or modified. It is worth exploring how to extend our approach to allow for examples with multiple sense labels and cases where one label takes precedence over another. We believe that this will be an important property of the work.

Another limitation is that we only use English datasets. There are PDTB-style datasets in other languages including a Chinese TED dicourse bank corpus \cite{Long2020TEDCDBAL}, a Turkish discourse Tree bank corpus \cite{zeyrek-kurfali-2017-tdb} and an Italian Discourse Treebank \cite{pareti-prodanof-2010-annotating}. Moreover, \citet{zeyrek2019ted} proposed a TED Multilingual Discourse Bank (TED-MDB) corpus, which has 6 languages. These datasets allow us to assess the approach in languages other than English. Besides, there are datasets similar to PDTB-Style like Prague Dependency Treebank \cite{Mrovsk2014DiscourseRI}. The different datasets use essentially similar sense hierarchy, but two things need to be investigated (i) whether there are comparable differences between tokens that realise “sister” relations, or (ii) whether tokens often have multiple sense labels, which would change what could be used as negative examples if leveraging our approach on them. 

In the future, we can also assess whether contrastive learning could help in separating out EntRel relations and AltLex relations from implicit relations or whether other methods would perform better.

% Regarding those instances with multiple annotated labels, following previous work \cite{qin-etal-2017-adversarial}, we treat the examples with multiple annotated labels as separate examples during training. At test time, a prediction matching one of the gold types is taken as the correct answer. Therefore, the approach might be inadequate for dealing with the actual distribution of the data and can be extended or modified. 

% Make the case that  What we don’t know is (i) whether there are comparable differences between tokens that realise “sister” relations, or (ii) whether tokens often have multiple sense labels, which would change what could be used as negative examples.  For instance if tokens with label L also have label M, and M is a sister or niece of L, then tokens labelled M would not be effective negative examples of tokens labelled L.  Also refer to the precedence of one label over another. So if tokens that satisfy both the criteria for sense L and for sense M must be labelled L, according to the annotation manual, again tokens labelled M would not be effective negative examples for L.

\section{Conclusions}

% We present a novel model, BMGF-RoBERTa, that combines representation, matching, and fusion modules for implicit discourse relation classification. Experimental results show BMGF-RoBERTa outperforms all previous state-of-the-art systems with substantial margins on the PDTB dataset and CoNLL datasets. We find the “previous-next” context re- lationship provides much information for this difficult task. Furthermore, the effectiveness of these well-designed mod- ules and what is important in the representation layer are thor- oughly analyzed in the ablation study. One important future work would be to find more clues for better representations within and even beyond the word- and sentence-levels.
In this paper, we leverage the sense hierarchy to select the negative examples needed for contrastive learning for the task of implicit discourse relation recognition. Our method has better overall performance than achieved by previous systems, and compared with previous work, our method is better at learning minority labels. Moreover, we compared different methods of selecting the negative examples based on the hierarchical structures, which shows some potential negative impacts might be produced when negative examples include those from other level-1 types. Moreover, we conduct ablation studies to investigate the effects of our data augmentation method and our contrastive learning method. Besides, the limitations and the future work are discussed.

\section*{Acknowledgments}
This work was supported in part by the UKRI Centre for Doctoral Training in Natural Language Processing, funded by the UKRI (grant EP/S022481/1), the University of Edinburgh. The authors also gratefully acknowledge University of Edinburgh Huawei Laboratory for their support. 

\bibliography{contrastive}
\bibliographystyle{acl_natbib}

\appendix
\clearpage

\section{Appendix}
% In our experiments for other negatives defining methods, we use the same hyperparameters as the experimental setup of our methods. The learning rate is $3e{-}5$ and the batch size is 256. For method 3 and method 4, the weight of positive examples is set to 1.6 and 1.3 and the weight of negative examples still is 1.
% %\label{sec:appendix}

\subsection{PDTB Hierarchy}
The hierarchies of both PDTB 2.0 and PDTB 3.0  consist of three levels, but for implicit relation recognition, so far no classification for third level labels has been done. We also focus on the hierarchy between level-1 and level 2. The PDTB-3 relation hierarchy simplifies and extends the PDTB-2 relation hierarchy. The PDTB 3.0 hierarchy not only simplifies the PDTB-2 relation hierarchy by restricting Level-3 relations to differences in directionality and eliminating rare and/or difficult-to-annotate senses, but also augments the relation hierarchy. Figure 6 and Figure 7 show PDTB 2.0 relation hierarchy and PDTB 3.0 relation hierarchy respectively.

\begin{figure}[htbp]
    \centering
    \includegraphics[width=8cm]{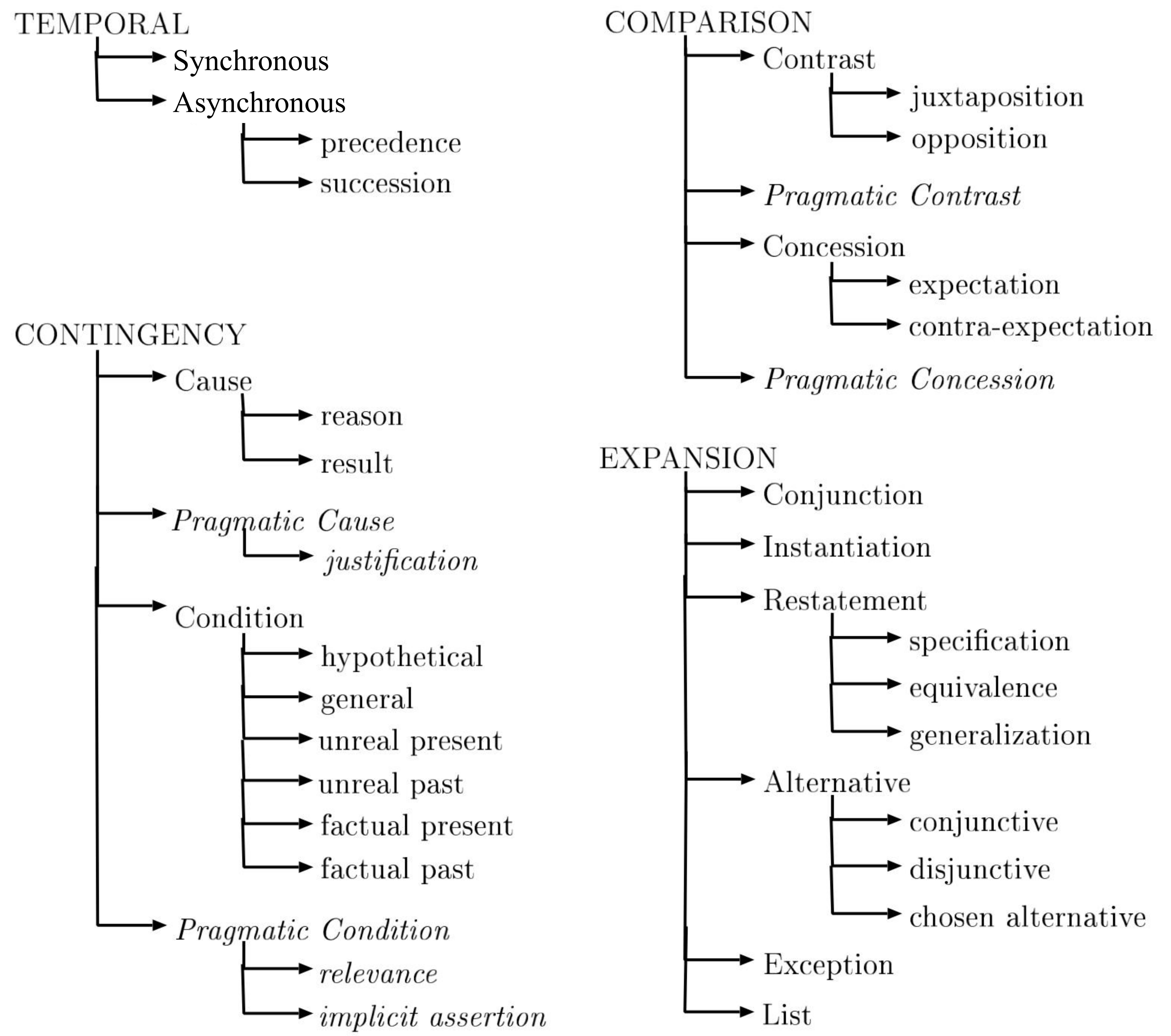}
    \caption{The PDTB 2.0 Senses Hierarchy.}
    \label{fig:my_label}
\end{figure}
\begin{figure}[htbp]
    \centering
    \includegraphics[width=8cm]{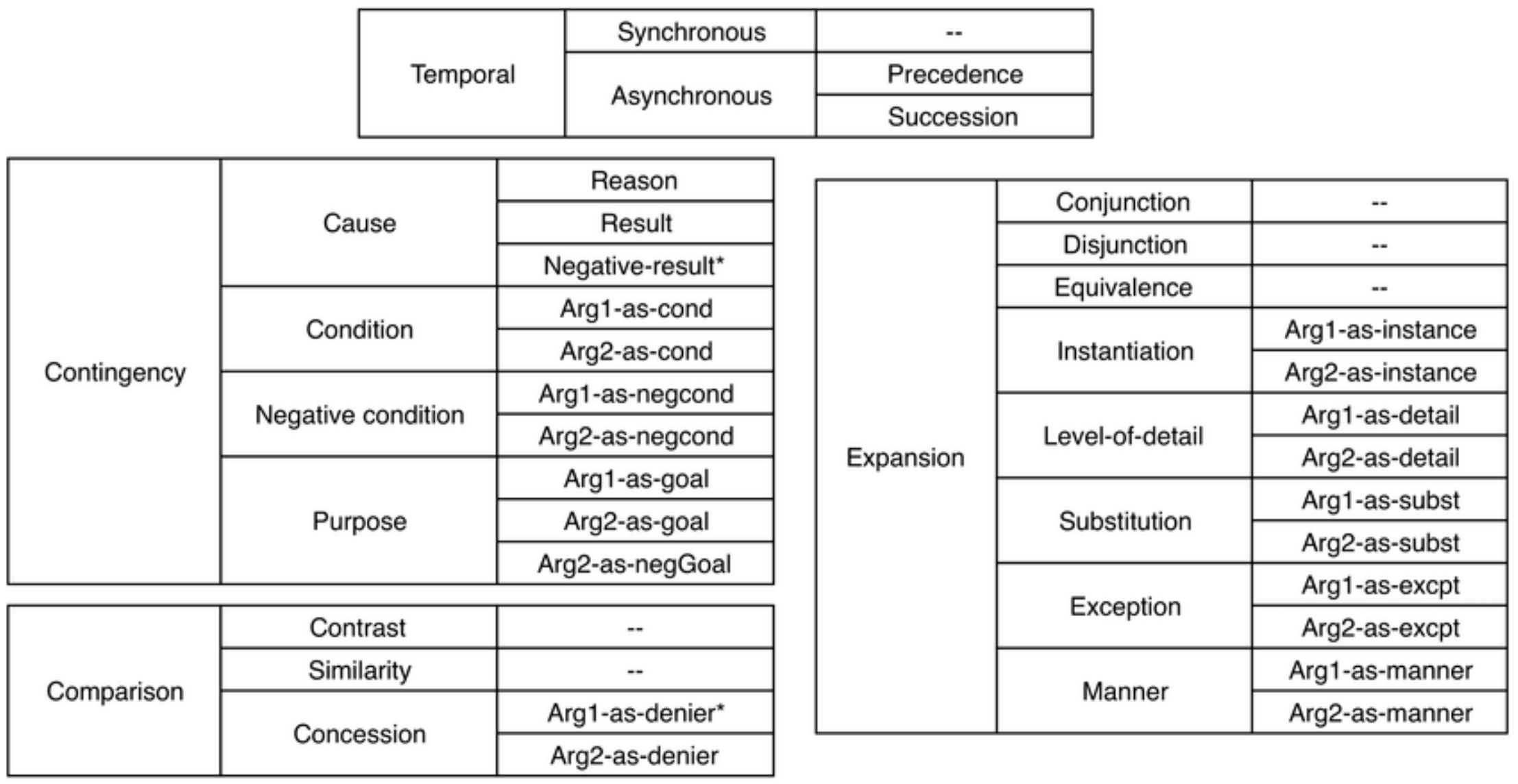}
    \caption{The PDTB 3.0 Senses Hierarchy. The leftmost column contains the Level-1 senses and the middle column, the Level-2 senses. For asymmetric relations, Level-3 senses are located in the rightmost column.}
    \label{fig:my_label}
\end{figure}
% The hierarchies of PDTB 2.0 and PDTB 3.0 all consists of three levels, but for implicit relation recognition, so far no classification for third level labels has been done. We also focus on the hierarchy between level-1 and level 2. The PDTB-3 relation hierarchy simplifies and extends the PDTB-2 relation hierarchy. PDTB 3.0 hierarchy not only simplify the PDTB-2 relation hierarchy by restricting Level-3 relations to differences in directionality and eliminating rare and/or difficult-to-annotate senses, but also augment the relation hierarchy. Figure 1 and Figure 2 show PDTB 2.0 relation hierarchy and PDTB 3.0 relation hierarchy respectively. 

% \subsection{Experiment Setting for other negatives defining methods}
% % \label{sec:appendix}
\label{sec:appendix}
\subsection{The results on relation types on PDTB-3}
We also examine the classification performance on PDTB-3 in terms of Macro-F1 for the four main relation types at level-1 and 14 sense types at level-2. The results can be seen in Table 9 and Table 10. Our model has significantly better performance for all level-1 relations. 

As for level-2 sense types, because there are no results of previous systems, we just show the result of 14 level-2 sense types in PDTB-3 in terms of F1. 

% \begin{figure}[htbp]
%     \centering
%     \includegraphics[width=8cm]{images/pdtb3.jpg}
%     \caption{The PDTB-3 Senses Hierarchy. The leftmost column contains the Level-1 senses and the middle column, the Level-2 senses. For asymmetric relations, Level-3 senses are located in the rightmost column.}
%     \label{fig:my_label}
% \end{figure}
\begin{table}[tbp]
\centering
\scriptsize
\begin{tabular}{l|cc|cc}
\bottomrule

Model&Comp.&Cont&Exp.&Temp.\\
\hline
\citet{liu-li-2016-recognizing}& 29.15&63.33&65.10& 41.03\\
\citet{lan-etal-2017-multi} &30.10&60.91&64.03& 33.71\\
\citet{ruan-etal-2020-interactively}& 30.37&61.95&64.28&  34.74\\
\citet{chen-etal-2016-implicit} & 27.34&62.56 &64.71&38.91\\
\citet{xiang-etal-2022-encoding}& 34.16 &65.48&67.82&  40.22\\
(BiLSTM)&&&&\\
\citet{xiang-etal-2022-encoding} & 35.83&66.77&70.00& 42.13\\
(BERT)&&&&\\

\hline
% RoBERTa&69.10&64.67&56.40&43.74\\
% RoBERTa-multi&67.37&62.04&55.53&36.85\\
% RoBERTa&61.39 &59.66 &75.75 &62.67 \\
% RoBERTa-MTL&60.84 &63.57 &77.31 &59.85 \\
% \hline
% $\rm RoBERTa_{base}$+$\rm SCL_{WL1}$&63.26 &60.42 &76.78 &59.74 \\
% $\rm RoBERTa_{base}$+$\rm SCL_{WL2}$&60.78 &60.82 &77.89 &56.30 \\

% $\rm RoBERTa_{base}$+$\rm SCL_{WL1+WL2}$&59.85 &65.18 &76.43 &64.67 \\
% $\rm RoBERTa_{base}$+$\rm SCL_{WL1+WL2+WL3}$&57.25&61.73&77.30 &64.90 \\
% \hline
Ours&\textbf{63.30}&\textbf{78.60}&\textbf{79.91}&\textbf{58.39}
\\
\hline
\toprule
\end{tabular}
\caption
{The results of different relations on PDTB-3 in terms of F1 (\%) (top-level multi-class classification).}
\label{table-1}
\end{table}

\begin{table}[!tbp]
\centering
\scriptsize
\begin{tabular}{l|c}
\bottomrule

Second-level Label&Ours\\
\hline
Temp.Asynchronous&66.35\\
Temp.Synchrony&41.38\\
\hline
Cont.Cause&71.38 \\
Cont.Cause+Belief&0.0\\
Cont.Condition&74.07\\
Cont.Purpose&96.05\\
\hline
Comp.Contrast &56.91 \\
Comp.Concession& 60.11 \\
\hline
Exp.Conjunction &61.70\\
Exp.Equivalence&11.43\\
Exp.Instantiation&69.83\\
Exp.Level-of-detail &55.34 \\
Exp.Manner  &78.43\\
Exp.Substitution &63.77 \\
% \citet{Guo2020WorkingMN}&43.92& 57.67 &73.45 &36.33\\
% \citet{Nguyen2019EmployingTC}&48.44& 56.84 &73.66 &38.60\\
% \citet{Liu2020OnTI}&59.44 &60.98 &77.66 & 50.26\\
% \citet{jiang-etal-2021-just}&55.40& 57.04 &74.76& 41.54\\
% \citet{dou-etal-2021-cvae-based}&55.72 &63.39 &80.34 &44.01\\
% \citet{DBLP:journals/corr/abs-2112-11740}&\\

% \hline
% % RoBERTa&69.10&64.67&56.40&43.74\\
% % RoBERTa-multi&67.37&62.04&55.53&36.85\\
% RoBERTa&61.39 &59.66 &75.75 &62.67 \\
% RoBERTa-MTL&60.84 &63.57 &77.31 &59.85 \\
% \hline
% $\rm RoBERTa_{base}$+$\rm SCL_{WL1}$&63.26 &60.42 &76.78 &59.74 \\
% $\rm RoBERTa_{base}$+$\rm SCL_{WL2}$&60.78 &60.82 &77.89 &56.30 \\

% $\rm RoBERTa_{base}$+$\rm SCL_{WL1+WL2}$&59.85 &65.18 &76.43 &64.67 \\
% $\rm RoBERTa_{base}$+$\rm SCL_{WL1+WL2+WL3}$&57.25&61.73&77.30 &64.90 \\
% \hline
% Ours&65.84 &63.55 &79.17 &69.86 
% \\
\hline
\toprule
\end{tabular}
\caption
{The results of different relations on PDTB-3 in terms of F1 (\%) (second-level multi-class classification).}
\label{table-4}
\end{table}

\end{document}